\documentclass[11pt]{article}
\usepackage[left=1in,right=1in,top=1in,bottom=1in]{geometry}
\usepackage{times}
\usepackage{hyperref}
\usepackage{graphicx}
\usepackage{amsmath}
\usepackage{listings}
\usepackage{xcolor}

\usepackage{wrapfig}

\usepackage{authblk}

\hypersetup{
  colorlinks   = true, 
  urlcolor     = blue, 
  linkcolor    = blue, 
  citecolor   = red 
}

\definecolor{codegreen}{rgb}{0,0.6,0}
\definecolor{codegray}{rgb}{0.5,0.5,0.5}
\definecolor{codepurple}{rgb}{0.58,0,0.82}
\definecolor{backcolour}{rgb}{0.95,0.95,0.92}
\definecolor{codered}{rgb}{0.5,0,0}

\lstdefinestyle{mystyle}{
    backgroundcolor=\color{backcolour},   
    commentstyle=\color{codered},
    keywordstyle=\color{magenta},
    numberstyle=\tiny\color{codegray},
    stringstyle=\color{codegreen},
    breakatwhitespace=false,         
    breaklines=true,                 
    captionpos=b,                    
    keepspaces=true,                 
    numbers=left,                    
    numbersep=5pt,                  
    showspaces=false,                
    showstringspaces=false,
    showtabs=false,                  
    tabsize=2,
    basicstyle=\ttfamily\footnotesize
}
 
\graphicspath{{images/}}

\begin{document}

\title{\huge PyRep: Bringing V-REP to \\ Deep Robot Learning}
\author[1]{Stephen James}
\author[2]{Marc Freese}
\author[1]{Andrew J. Davison}
\affil[1]{Department of Computing, Imperial College London, UK}
\affil[2]{Coppelia Robotics, Switzerland}
\date{}

\maketitle

\begin{abstract}
PyRep\footnote{\href{https://github.com/stepjam/PyRep}{https://github.com/stepjam/PyRep}} is a toolkit for robot learning research, built on top of the virtual robotics experimentation platform (V-REP). Through a series of modifications and additions, we have created a tailored version of V-REP built with robot learning in mind. The new PyRep toolkit offers three improvements: (1) a simple and flexible API for robot control and scene manipulation, (2) a new rendering engine, and (3) speed boosts upwards of $10,000\times$ in comparison to the previous Python Remote API. With these improvements, we believe PyRep is the ideal toolkit to facilitate rapid prototyping of learning algorithms in the areas of reinforcement learning, imitation learning, state estimation, mapping, and computer vision.
\end{abstract}

\section{Introduction}

In recent years, deep learning has significantly impacted numerous areas in machine learning, improving state-of-the-art results in tasks such as image recognition, speech recognition, and language translation~\cite{lecun2015deep}. Robotics has benefited greatly from this progress, with many robotics systems opting to use deep learning in many or all of the processing stages of a typical robotics pipeline~\cite{zeng2018robotic, morrison2018cartman}. As we aim to endow robots with the ability to operate in complex and dynamic worlds, it becomes important to collect a rich variety of data of robots acting in these worlds. If we are to use deep learning however, it comes at a cost of requiring large amounts of training data, which can be particularly time consuming to collect in these dynamic environments. Simulations then, can help in one of two primary ways:

\begin{itemize}
    \item Rapid prototyping of learning algorithms in the hope to find data-efficient solutions that can be trained on small real-world datasets that are feasible to collect.
    \item Training on a large amount of simulation data with potentially a small amount of real-world data, and find ways of transferring this knowledge from simulation to the real world~\cite{bousmalis2018using, james2018sim, tobin2017domain, james2017transferring, matas2018sim}.
\end{itemize}

Two common simulation environments in the literature today are Bullet~\cite{bullet} and MuJoCo~\cite{Todorov2012MuJoCoAP}. However, given that these are physics engines rather than robotics frameworks, it can often be cumbersome to build rich environments and integrate standard robotics tooling such as inverse \& forward kinematics, user interfaces, motion libraries, and path planners.

Fortunately, the Virtual Robot Experimentation Platform (V-REP)~\cite{rohmer2013v} is a robotics framework that makes it easy to design robotics applications. However, although the platform is highly customisable and ships with several API, including a Python remote API, it was not developed with the intention to be used for large-scale data collection. As a result, V-REP, when accessed via Python,  is currently too slow for the rapid environment interaction that is needed in many robot learning methods, such as reinforcement learning (RL). To that end, PyRep is an attempt to bring the power of V-REP to the robot learning community. In addition to a new intuitive Python API and rendering engine, we modify the open-source version of V-REP to tailor it towards communicating with Python; as a result, we achieve speed boosts upwards of $10,000 \times$ in comparison of the original V-REP Python API.

\section{Background}

V-REP~\cite{rohmer2013v} is a general-purpose robot simulation framework maintained by \textit{Coppelia Robotics}. Some of its many features include:
\begin{itemize}
    \item Cross-platform content (Linux, Mac, and Windows).
    \item Several means of communication with the framework (including embedded Lua scripts, C++ plugins, remote APIs in 6 languages, ROS, etc).
    \item Support for 4 physics engines (Bullet, ODE, Newton, and Vortex), with the ability to quickly switch from one engine to the other. 
    \item Inverse \& forward kinematics.
    \item Motion planning.
    \item Distributed control architecture based on embedded Lua scripts.
\end{itemize}

Python and C++ are primary languages for research in deep learning and robotics, and so it is imperative that communication times between a learning framework and V-REP are kept to a minimum. Given that V-REP was introduced in 2013 when deep learning was in its infancy, prioritisation was not given to rapid external API calls, which currently rely on inter-thread communication. As a result, this makes V-REP slow to use for external data-hungry applications.

\section{Modifications}

Below we outline the modifications that were made to V-REP.

\paragraph{Speed.} The 6 remote APIs offered suffer from 2 communication delays. One of these comes from the socket communication between the remote API and the simulation environment (though this can be decreased considerably using shared memory). The second delay, and most notable, is inter-thread communication between the main thread and the various communication threads. This communication latency can become noticeable when the environment needs to be queried synchronously at each timestep (which is often the case in RL). To remove these latencies, we have modified the open-source version of V-REP such that Python now has direct control of the simulation loop, meaning that commands sent from Python are directly executed on the same thread. With these modifications we were able to collect robot trajectories/episodes over 4 orders of magnitude faster than using the original remote Python API; making PyRep an attractive platform for evaluation of robot learning methods.

\begin{figure}
  \centering
      \includegraphics[height=0.22\linewidth]{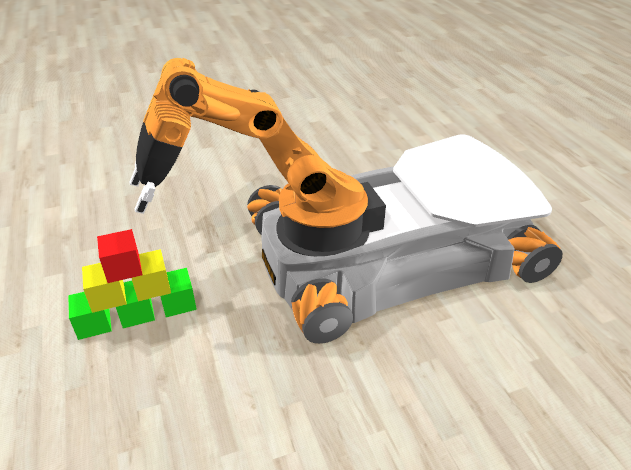}    
      \includegraphics[height=0.22\linewidth]{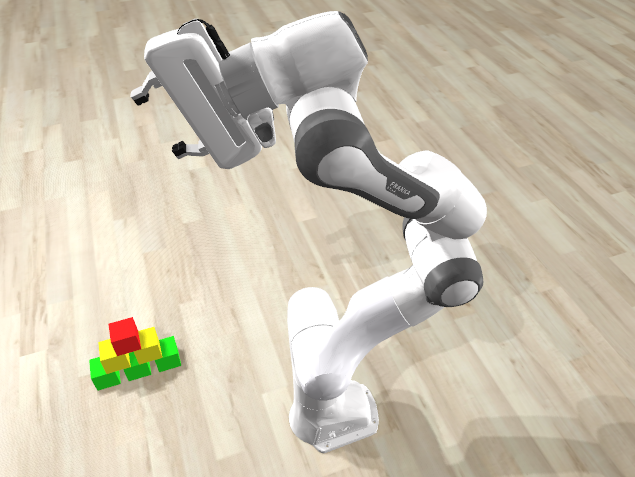}        
      \includegraphics[height=0.22\linewidth]{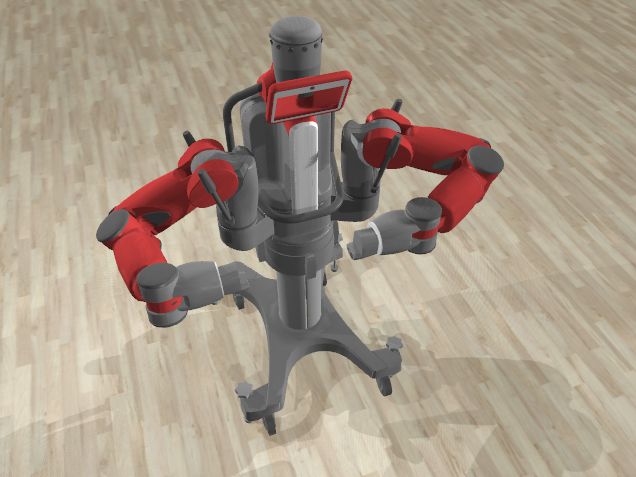}       
  \caption{Example images of environments using the new renderer.}
  \label{fig:envs_and_shadows}
\end{figure}

\paragraph{Renderer.} V-REP ships with 2 main renderers: a default OpenGL 2.0 renderer, and the POV-Ray ray tracing renderer. POV-Ray produces high quality images but at a very low framerate. The OpenGL 2.0 renderer on the other hand uses basic shadow-free rendering, and uses the old-style fixed-function pipeline OpenGL. As part of this report, we release a new OpenGL 3.0+ renderer which supports shadow rendering from all V-REP supported lights, including directional, spotlight, and pointlight. Examples renderings can be seen in Figure \ref{fig:envs_and_shadows}.

\paragraph{PyRep API.} The new API manages simulation handles and provides an object-oriented way of interfacing with the simulation environment. Moreover, we have made it easy to add new robots with motion planning capabilities with only a few lines of Python code. An example of the API in use can be seen in Figure \ref{fig:api_example}. 

\begin{figure}[h]
\begin{lstlisting}[language=Python]
 from pyrep import PyRep
 from pyrep.objects import VisionSensor, Shape
 from pyrep.arms import Franka

 pr = PyRep()
 pr.launch('my_scene.ttt', headless=True)   # Launch V-REP in a headless window
 pr.start()  # Start the physics simulation

# Grab robot and scene objects
 franka = Franka()
 camera = VisionSensor('my_camera')
 target = Shape('target')

 while training:
     target.set_position(np.random.uniform(-1.0, 1.0, size=3))  
     episode_done = False
     while not episode_done:
         # Capture observations from the vision sensor
         rgb_obs = camera.capture_rgb()
         depth_obs = camera.capture_depth()
         action = agent.act([rgb_obs, depth_obs])  # Neural network predicting actions
         franka.set_target_joint_velocities(action)  # Send actions to the robot
         pr.step()  # Step the physics simulation
         # Check if the agent has reached the target
         episode_done = target.get_position() == franka.get_tip().get_position()
\end{lstlisting}
\caption {PyRep API Example. Many more examples can be seen on the GitHub page.}
\label{fig:api_example}
\end{figure}

\section{Conclusion}

V-REP has been used extensively over the years in more traditional robotics research and development, but has been overlooked by the growing robot learning community. The new PyRep toolkit brings the power of V-REP to the community by providing a simple and flexible API, significant speedup in run-time, and integration of an OpenGL 3.0+ renderer to V-REP. We are eager to see the tasks that can be solved by new and exciting robot learning methods.

\bibliographystyle{ieeetr}
\bibliography{mybib}

\end{document}